# Handwritten Word Recognition using Deep Learning Approach: A Novel Way of Generating Handwritten Words

Mst Shapna Akter*, Hossain Shahriar†, Alfredo Cuzzocrea‡, Nova Ahmed §, Carson Leung¶

*Department of Computer Science, Kennesaw State University, USA
† Department of Information Technology, Kennesaw State University, USA
‡ iDEA Lab, University of Calabria, Rende, Italy
§ Department of Electrical and Computer Engineering (ECE), North South University, Bangladesh
¶ Department of Computer Science, University of Manitoba, Canada

{*makter2@students.kennesaw.edu, † hshahria@kennesaw.edu, ‡ alfredo.cuzzocrea@unical.it, § nova.ahmed@northsouth.edu, , ¶ kleung@cs.umanitoba.ca }

*Abstract*—A handwritten word recognition system comes with issues such as– lack of large and diverse datasets. It is necessary to resolve such issues since millions of official documents can be digitized by training deep learning models using a large and diverse dataset. Due to the lack of data availability, the trained model does not give the expected result. Thus, it has a high chance of showing poor results. This paper proposes a novel way of generating diverse handwritten word images using handwritten characters. The idea of our project is to train the BiLSTM-CTC architecture with generated synthetic handwritten words. The whole approach shows the process of generating two types of large and diverse handwritten word datasets: overlapped and non-overlapped. Since handwritten words also have issues like overlapping between two characters, we have tried to put it into our experimental part. We have also demonstrated the process of recognizing handwritten documents using the deep learning model. For the experiments, we have targeted the Bangla language, which lacks the handwritten word dataset, and can be followed for any language. Our approach is less complex and less costly than traditional GAN models. Finally, we have evaluated our model using Word Error Rate (WER), accuracy, f1-score, precision, and recall metrics. The model gives 39% WER score, 92% percent accuracy, and 92% percent f1 scores using non-overlapped data and 63% percent WER score, 83% percent accuracy, and 85% percent f1 scores using overlapped data.

## I. INTRODUCTION

Handwritten words, handwritten texts, and handwritten character recognition are becoming popular for investigation. Handwritten word Recognition is the procedure of automatic conversion of a handwritten word into a machine-encoded word [1]. The terms "offline HWR" and "online HWR" refer to two distinct issues that are covered by handwritten word recognition. Offline HWR addresses the issue of reading a handwritten word offline, or at a later time (minutes, months, or years) than when it was originally written [2]. A handwritten word is often entered into a recognition algorithm as a binary or grayscale picture after being scanned from a paper document. In contrast to online HWR, when writing is done with a special pen on a tablet or an electronic notepad and the recognition algorithm has access to temporal information like the pen's position and velocity along its trajectory, the challenge involves writing with a regular pen on a regular paper [2]. Online HWR is also called real-time HWR because most online HWR algorithms try to recognize the writing as it is being produced. This research focuses on automatically identifying the offline handwritten word. However, automatically reading handwritten texts is a challenging task because of the many differences in character shape, as well as the overlapping and interconnectivity of the nearby characters. Characters are frequently unclear when seen alone. Therefore, context is crucial to reduce classification mistakes. The current development efforts have encompassed extensive evolutions of many categorization algorithms, typically yielding a final design that is a technical synthesis of many different methodologies. Handwritten character recognition is common, and many investigations have been done in many countries, including Bangladesh [3–6]. From the previous work, we found that segmentation-based word recognition assigns character confidence levels for the segments of words to account for the ambiguity within character classes [7]. Numerous design initiatives for character recognition are theoretically based on practically all classification techniques, including neural nets, linear discriminant functions, fuzzy logic [8], template matching [9], binary comparisons, etc [10]. It is equally crucial to choose the nominal method to use as it is to decide which attributes to analyze and how to measure them. When considering upper and lower case characters as separate classes, there are as many as 52 classes, with varying degrees of difficulty for recognition depending on the complexity, style differences, and ambiguity among them. The variances between upper and lower case, cursive versions of each letter, and the sheer number of multi-stroke characters that are intrinsically broken make isolated character recognition much more difficult. The collection of sufficient samples under realistic circumstances plays a crucial role in the final success of an attempt. The distribution of samples among the various classes in the training set is typically not uniform and may also follow a different distribution from the expected occurrence or relative, which is another issue with character recognition. Therefore, computerized word recognition system seems to be easier and more efficient if analyzed with large-scale datasets. Bangladesh lacks a large and diverse handwritten word dataset. The handwritten character dataset is available on the BanglaAi.com website, which can only be used for the Bangla optical recognition system. Therefore, it is essential to create a handwritten word dataset. Previously, some of the papers focused on generating synthetic handwritten word datasets using a generative adversarial network (GAN) [11]. GAN nowadays plays an important role in generating images with different purposes. However, GAN lacks to mimic the diverse range of handwritten words, and it requires a dataset for training before generating the new dataset [12]. Since Bangla completely lacks the handwritten word dataset, it is not even possible to train the GAN network with the existing dataset. Moreover, generating the dataset using the GAN network is quite costlier and more complex.

In our paper, we have overcome all of the issues by using a very simple method. We have generated the handwritten word dataset using the handwritten character by joining the horizontal lines by following the Bangla dictionary. In addition, we have created an overlapped dataset following the same but not an identical process. The variation shows overlapping between the characters, spelling mistakes, cursive nature, etc. We have tried to put the variation in writing style by following different grapheme combinations while joining them. In the handwritten word, many graphemes overlap within the word. We have created overlapped data by overlapping 4 pixels from the left image. Our dataset produces diverse and meaningful words which can be used for training the network. Before generating the handwritten word, we have done some basic preprocess techniques such as resized and converted images into greyscale, fed the words and their corresponding labels to the BiLSTM-CTC network and evaluated the network using WER, accuracy, F1-score, precision, and recall. This approach would contribute to recognizing handwritten words for languages left behind due to the unavailability of the dataset. Therefore, the whole process can be followed for any language which has handwritten characters available.

The rest of this paper is arranged as follows. Section 2 provides the background needed for the study. The data sources, preprocessing methods, and models used in this work for aggression detection tasks are discussed in Section 3. The simulation results based on the classification algorithms are analyzed in Section 4. Section 5 provides the overview of qualitative study. Finally, this paper is summarized in Section 6.

## II. LITERATURE REVIEW

Previously, many investigations have been done on the OCR field as it brings many challenges, such as variation in writing, variation in style, poor quality of the source, and so on [13–17]. Bangla is one of the cursive nature languages, which comes with another challenge. From the previous investigations, segmentation, holistic, HMM, Neural network model, and LSTM-CTC based network have been found for Handwriting recognition in OCR. Apart from these techniques, some literature has also shown performance improvement-based approaches. We have tried to go through the related papers on optical character recognition (OCR). Alam et al. [13] showed a complete OCR system for printed character recognition. They preprocessed a dataset using text digitization, binarization, noise removal, skew detection and correction, and segmentation techniques. They also performed the line-level, word-level, and character-level segmentation processes. Here the upper zone denotes the portion above the headline, the middle zone denotes the character, and the lower zone denotes the modifiers. They removed the headline and scanned vertically using a piecewise linear approach. After getting the whole character, they extracted the feature from the character and classified it using a neural network. Their methodology shows almost 97% accuracy. A simple line segmentation process is shown in the paper of Malik et al. [14], for Urdu handwritten and printed text. The technique depends on counting continuous black pixels along the horizontal line. Their approach also worked on the Skew line. Their training dataset consisted of 80 pages of handwritten Urdu text and 48 pages of printed text lines. The performance showed good in printed documents but poorly worked on overlapping handwritten words. A different segmentation process is shown in Pastor et al. [18], who presented a Convolutional Neural Network-based method for the extraction of text lines. The computation was done on the main body area. A region-based method using a watershed transform was performed on the main body area to extract the resulting lines. The model requires a reliable ground truth or labeled data, which is difficult to handle. Sahoo et al. [19] followed a holistic word recognition approach to recognize handwritten Bangla words. their dataset consists of 12000 handwritten Bangla words. They used 80 classes which consisted of 150 samples per class. For preprocessing, they used otsu's binarization process; for feature extraction, they adopted negative refraction of light. Basically, they proposed a feature vector extracted by transforming word images into VCC. However, the feature vector's size increases the computational time and low robustness due to the word-class limitations. They achieved 87.50 % accuracy using their approach. Malakar et al. [20] used the similar way to segment Hindi words. They collected handwritten Hindi data and extracted the features. They extracted each word's low-level features such as area, aspect ratio, pixel density, pixel ratio, longest run, centroid, and projection length. To evaluate the recognition performance, they classified the feature vector using five classifiers such as MLP, SMO, LRM, Naive Bayes, and Multiclass. They achieved the best accuracy of 96.82 % using the MLP classifier. O. Samantha et al. [21] segmented the handwritten word samples. The recognition step is done by using the HMM method.

Similiar approach proposed in Sen et al. [22]. Their word recognition module comprises different modules: preprocessing the word samples, segmentation of words into basic strokes, recognizing the basic strokes using multilayer perceptron, and recognition process used by the Hidden Markov Model (HMM). A total of 50 different word samples with 110 instances were used to evaluate the proposed model. They obtained 95.4% and 90.3% accuracy in stroke-level and word-level data. Basically, HMM avoids character-level segmentation. Most of the segmentation process requires character-level segmentation to segment words, where HMM shows an advantage in this case. Some simple Neural Network and Convolutional Neural network-based investigation has been done in this field, which shows satisfactory results in some cases. Very high accuracy is shown in Rabby et al. [23], where they used a simple CNN model to classify Bangla handwritten characters. The proposed model is a 13-layer convolutional neural network with two sub-layers. They performed data augmentation to avoid the overfitting problem and got the highest accuracy of 98%. Compound character recognition using a neural network is proposed in Fardous et al. [24]. They performed both the convolutional and fully connected layers to train their model. The experiment result shows that the network highly overfits the training data as the training data is minimal to train in the deep network. So, they performed augmentation using some noises and increased the training data to overcome the issue. The result of the network showed 95.5% accuracy on the test dataset. A simple Neural Network has some limitations while using a long sequence of input. LSTM combined with CTC loss function has become popular in this field. Here, the advantage is a long sequence of input can be trained using the network without any issue, and The CTC loss function performs best to filter the output sequence. Breuel et al. [25] introduced a line recognizer using a combination of deep convolutional networks and LSTM. Their novelty was brought by the combination of convolutional layers, max-pooling, and 1D LSTM, which showed a lower error rate than the older 1D LSTM network. Another similar approach is shown in Sabbir et al. [26], who used a segmentation-free model, and the LSTM output requires only decoding. Later the CTC loss function and Character Error Rate (CER) for error measurement was used. Rawls et al. [27] proposed a simple and effective LSTM-based approach to recognize machine-print text from raw pixels. During recognition, they trained the network using the CTC objective function and a WFST language model. Biswas et al. [28] have considered the line-level recognition of degraded printed Bangla documents. The Gaussian mixture model is used to extract texts from the noisy background of degraded documents. They used a hybrid neural network architecture consisting of a convolutional neural network (CNN), two layers of BLSTM cells, and a connectionist temporal classification (CTC) to recognize the segmented text line. There exist many limitations with the approaches such as segmentation, Holistic, Hidden Markov model (HMM), and simple NN and CNN model. We have ruled out those techniques from our work. However, due to the performance and efficiency, we have been inspired by some of the previous work related to LSTM-CTC architecture.

## III. METHODS

We have taken meaningful Bangla words from the Bangla dictionary wordnet and joined a diverse set of handwritten character images from Bangla.Ai to resemble almost real handwritten word images. This way, we have generated almost 60470 realistic word images for both the overlapped and non-overlapped datasets. The synthetic images are then fed to the CNN model to extract the features. The BiLSTM-CTC is one of the most popular deep learning architecture to recognize handwritten words. The whole training process is carried out with synthetic generated handwritten words. we have trained our model with 60470 data. The training data is split into two parts: 80 percent for training data and 20 percent for validation data. The performance of the model has been evaluated on test dataset.

All experiments are carried out on a Windows system with Intel (R) Core (TM) i7-1065G7 CPU @1.50GHz processor, 1TB HDD, 8GB RAM, a CUDA-enabled Nvidia G-Force MX250 4GB graphical processing unit (GPU). The codes are implemented with the Pytorch backend.

### A. Datasets

### B. bangla.Ai Handwritten character dataset

To generate the synthetic handwritten words, We have collected the handwritten Bangla graphemes from Bangla.AI . Grapheme refers to the set of units of a writing system, such as letters and letter combinations that represent a phoneme. In contrast, phonemes refer to any of the perceptually distinct units of sound in a specified language that distinguishes one word from another. The grapheme dataset contains 10,000 possible graphemes combined with three classes: grapheme root, vowel diacritics, and consonant diacritics. Bangla language has 49 letters in which 11 vowels and 38 consonants in its alphabet; there are also 18 potential diacritics or accents. The dataset contains the image of handwritten Bengali graphemes, useful for handwritten character recognition and developing handwritten words.

### C. Bangla word Dictionary

To generate above 64 thousand unique words, we have used BNNet (Bangla wordnet), a Bangla dictionary containing meaningful Bangla words. They have combined freely available online materials to build comprehensive Bangla word resources. It has 247748 unique Bangla words [29].

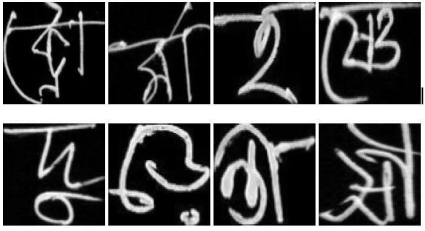

Fig. 1: Sample character images from Bangla.AI Handwritten character dataset

*D. Test Dataset*

BanglaWriting contains single-page handwritings of 260 individual people with different personalities and ages. Each of the pages bounds each word following bounding-box procedure. The dataset contains a total of 21234 words in which 5,470 words are unique. The bounding boxes and labeling of words are manually done by human. The dataset is useful for word recognition, writer identification, handwritten word segmentation, and word generation. We tested our models performance using the datatset [30].

*E. Data preprocessing*

*1) Training data:* Before generating the handwritten word, it requires to preprocess the datasets downloaded from the organization, since the images were not suitable to join due to the random size and extra space present in the image. The dataset is stored in parquet format. We have extracted the images from the files. All of the images have random space, which has been removed by using the tight bounding cropping process. the tight bounding cropping process crop the spaces until it finds white pixel from left, right, top and bottom. The tight bounding cropping removes the space between the images and makes it a realistic handwritten word. The cropped images have been resized into 128 X 128. The greyscale has been performed on all of the images.

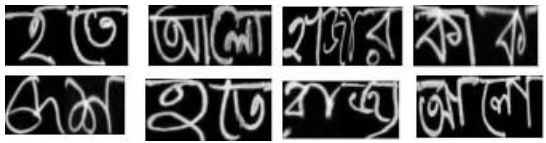

Fig. 2: Synthetically generated non-overlapped handwritten word images

*2) Test data:* The test dataset requires preprocessing before being fed into the model since it was in a json format for evaluation. To fed the data to the model, it was required to download as an image file. The images are then resized into 128 X 128 and performed greyscaling.

*F. Synthetic Dataset Generation*

Using random and meaningless words does not perform well in recognizing handwritten words. We ensure to generate meaningful words using the Bangla Dictionary. After finishing the preprocessing steps, the characters are picked according to the requirements to make meaningful words. We have used characters with a modifier or without a modifier and generated the synthetic data. The datasets we have generated are normally horizontal connected words and overlapping horizontal connected words. Normal horizontally connection refers to joining the horizontal pixels of the left image with the right image. On the other hand, the overlapping horizontal connection refers to joining the -4 pixels of the left image with the right image. The pixels may vary according to the requirements. In our case, we chose four pixels for generating the overlapped images. All words are meaningful. The graphical overview of the process has been shown in figure 5.

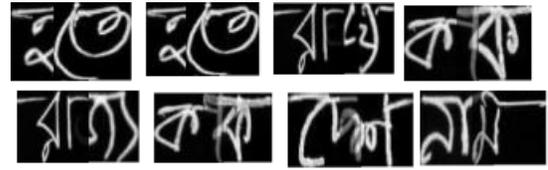

Fig. 3: Synthetically generated overlapped handwritten word images

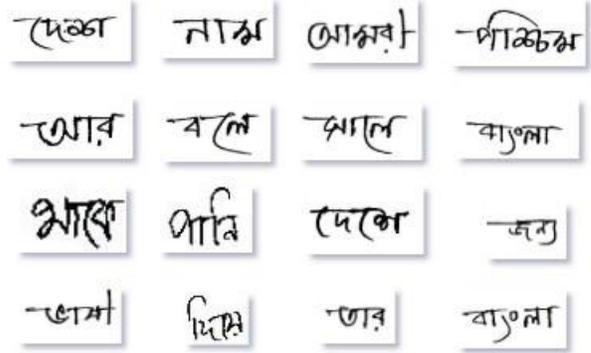

Fig. 4: Words extracted from json file

*G. Architecture*

The proposed architecture consists of convolution layers, followed by a convolution layer proceeded by bi-directional LSTM. The architecture finally ends with CTC loss function. The CTC loss function will direct the NN-training. We only provide the CTC loss function with the NN's output matrix and the matching ground-truth (GT) text. We fed an input size of 32X128 images to the CNN model. The CNN model extracts the features from the images; Then, the extracted features are fed to the Bi-directional long short-term memory (Bi-LSTM), which reads the features from left to right and generates a sequence. The sequence contains repetition of characters and blacks. CTC-loss is used to remove the repeating characters and blanks. BiLSTM uses a similar method to LSTM, with the exception that it permits backward propagation as well as forward propagation.

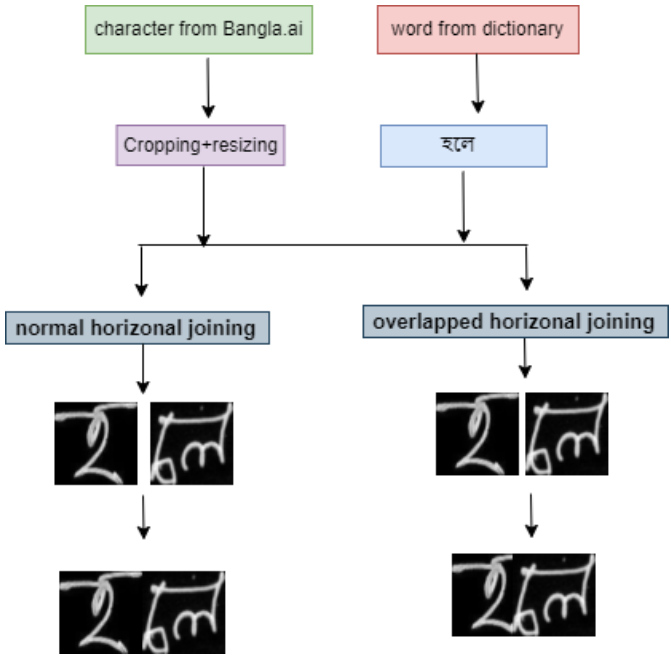

Fig. 5: A graphical overview of the process of generating the synthetic word

*1) LSTM:* LSTM is A popular artificial Recurrent Neural Network (RNN) model, which works with datasets that preserve sequences such as text, time series, video, and speech. Using this model for sequential datasets is effectiveSince as it can handle single data points. It follows the Simple RNN model's design and an extended version of that model. However, unlike Simple RNN, it has the capability of memorizing prior value points since it developed for recognizing long-term dependencies in the text dataset. Three layers make up an RNN architecture: input layer, hidden layer, and output layer [6, 31]. Figure 6 depicts the LSTM's structural layout. The elementary state of RNN architecture is shown as the mathematical function:

$$h_t = f(h_{t-1}, x_t; \theta) \quad (1)$$

Here, $\theta$ denotes the function parameter, $h_t$ denotes the current hidden state, $x_t$ denotes the current input, $f$ denotes the function of previous hidden state.

When compared to a very large dataset, the RNN architecture's weakness is the tendency to forget data items that are either necessary or unneeded. Due to the nature of time-series data, there is a long-term dependency between the current data and the preceding data. The LSTM model has been specifically developed to address this kind of difficulty. It is first proposed by Hochreiter Long [32].
This model's main contribution is its ability to retain long-term dependency data by erasing redundant data and remembering crucial data at each update step of gradient descent [33]. The LSTM architecture contains four parts: a cell, an input gate, an output gate, and a forget cell [34].

The purpose of the forget cell is to eliminate extraneous data by determining which data should be eliminated based on the state (t) Land input x(t) at the state c(t) 1.

At each cell state, the sigmoid function of the forget gate retains all 1s that are deemed necessary values and eliminates all 0s that are deemed superfluous.[32, 35–37]. The forget gate state's equation is stated as follows:

$$f_t = \sigma(W_f \cdot [h_{t-1}, x_t] + b_f) \quad (2)$$

where $f_t$ denotes sigmoid activation function, $h_{t-1}$ denotes output from previous hidden state, $W_f$ denotes weights of forget gate, $b_f$ denotes bias of forgetting gate function, and finally $x_t$ dentoes current input.

After erasing the unneeded value, new values are updated in the cell state. Three steps make up the procedure: The first step is deciding which values need to update using sigmoid layer called the "input gate layer". Second, creating a vector of new candidate values using the tanh layer. Finally, steps 1 and 2 are combined together to create and update the state.

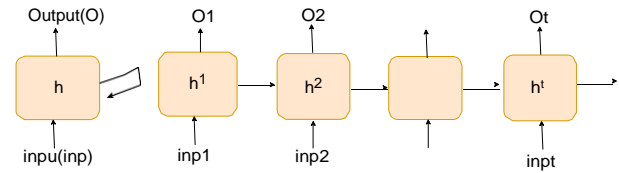

Fig. 6: LSTM neural network structure.

The equation for the sigmoid layer is as follows:

$$i_t = \sigma(W_i \cdot [h_{t-1}, x_t] + b_i) \quad (3)$$

The tanh layer generates a vector of new candidates for producing a new value to the state $C(t)$. The sigmoid layer determines which value should be updated.

The equation for tanh layer's equation is as follows:

$$\tilde{C}(t) = \tanh(W_c \cdot [h_{t-1}, x_t] + b_C) \quad (4)$$

The addition of $\tilde{C}(t) \cdot i_t$ and $C_t - 1 \cdot f_t$ updates the new cell at state $C(t)$. The updated state's equation is:

$$C_t = C_{t-1} * f_t + \tilde{C}(t) * i_t \quad (5)$$

In order to determine which output needs to be maintained, the output is ultimately filtered out using the sigmoid and the tanh functions.

$$O_t = \sigma(W_o \cdot [h_{t-1}, x_t] + b_o) \quad (6)$$

$$h_t = O_t * \tanh(C_t) \quad (7)$$

In this state, $h_t$ gives outputs that are used for the input of the next hidden layer.

*2) BiLSTM:* The Bidirectional Long short-term memory (BiLSTM) was first proposed by GRAVES [38]. The architecture of BiLSTM can learn patterns from both past-to-future and future-to-past data. This idea sets it apart from the LSTM model, which can learn patterns from the past to the future. Figure 7 depicts the bidirectional LSTM's structural layout. The backward propagation layer primarily functions as a forwarding LSTM reverse layer. The hidden layer synthesizes information from both the forward and backward directions [39]. As a result, "the reverse direction of forwarding direction" is used to calculate the LSTM reverse layer. The formula for computing the BiLSTM network is:

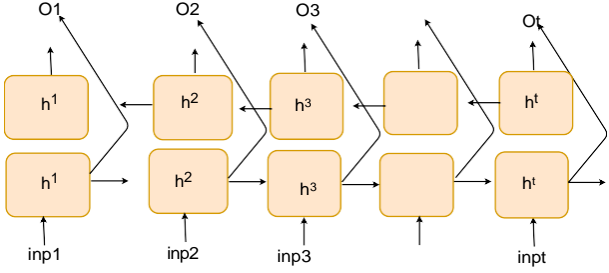

Fig. 7: BiLSTM neural network structure.

The mathematical equation [39] of backward propagation is as follows:

$$h_f = f(w_{f1}x_t + w_{f2}h_{t-1}) \qquad (8)$$

$$h_b = f(w_{b1}x_t + w_{b2}h_{t+1}) \qquad (9)$$

Where, $h_f$ denotes as the forward layer's output, and $h_b$ denotes as the reverse layer output.
The output derived from the hidden layer is given below:

$$O_i = g(w_{o1} * h_f + w_{o2} * h_b) \qquad (10)$$

*3) Connectionist Temporal Classification Loss (CTC):* CTC loss is developed for tasks that deal with alignment between sequences. It determines the loss between a continuous time series and a desired sequence. This is accomplished by adding up the likelihood of potential input-to-target alignments, which results in a loss value that may be differentiated with regard to each input node. Assuming a "many-to-one" alignment between the input and the target, the target sequence's length is constrained to be less than the input length [40].

*H. Evaluation metrics*

The evaluation metrics are used to evaluate a model's performance to check how close the predicted outputs are to the corresponding expected outputs. We have used Word Error Rate (WER), F1-score, precision, recall, and accuracy [41] as the evaluation metrics.

WER is mostly used to check a the speech recognition's models accuracy. It works by calculating the distance between the predicted result and the ground truth. we used Levenshtein distance for word level WER calculation.

$$WER = \frac{D(P, G)}{N} \qquad (11)$$

Where P refers to the predicted word sequence, G refers to the groundtruth word sequence, N refers to the number of words in the ground truth word sequence, and D(P,G) refers to the Levenshtein distance between P and G. However, we have also reported precision, recall, and f1-score for further comparison. Precision and recall are used to report the false positive and true negative values. Whereas, F1-score gives the average of both values.

**Precision :** precision calculates the number of correct values. In the word recognition classification, if the model gives low precision then many comments are incorrect, for high precision it will ignore the False positive values by learning with flase alarms [42]. The precision can be calculated as follows:

$$Precision = \frac{TP}{TP + FP} \qquad (12)$$

**Recall :**
Recall is opposite of Precision . Precision used when the false Nagatives are high [43].The recall can be calculated as follows:

$$Recall = \frac{TP}{TP + FN} \qquad (13)$$

**F1 score:** F1 score combines precision and recall and provides an overall measure of the models' accuracy. If the predicted value matches with the expected value, then the f1 score gives 1, and if none of the values matches with the expected value, it gives 0. [44]
The F1 score can be calculated as follows:

$$F1 = \frac{2 \cdot precision \cdot recall}{precision + recall} \qquad (14)$$

**Accuracy** : Accuracy determines how close the predicted output is to the actual value.

$$Accuracy = \frac{TP + TN}{TP + TN + FP + FN} \qquad (15)$$

here, TN refers to True Negative and FN refers to False Negative.

## IV. RESULTS

The evaluation process has been performed on test images that are unseen for the trained models. For Non-overlapped data, We have achieved 92 % accuracy and 39% percent of the total word error rate on the test dataset and out of the

21234 words, about 19507 words matched accurately. In case of overlapped data, we have achieved 83% accuracy and 93% of the total word error rate on the test dataset and out of the 21234 words, about 9112 words matched accurately.

TABLE-1 : Label distribution of dataset for subtask A

| Dataset | WER score | Accuracy | Precision | Recall | F1-score | Matched words |
|---|---|---|---|---|---|---|
| Non-overlapped | 0.39 | 0.92 | 0.87 | 0.92 | 0.92 | 19507 |
| Overlapped | 0.63 | 0.83 | 0.77 | 0.81 | 0.85 | 9112 |

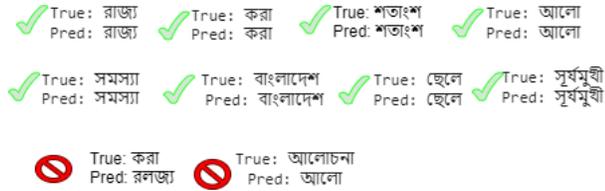

Fig. 8: Prediction on test dataset from the model trained with non-overlapped dataset

result of overlapped tranied model:

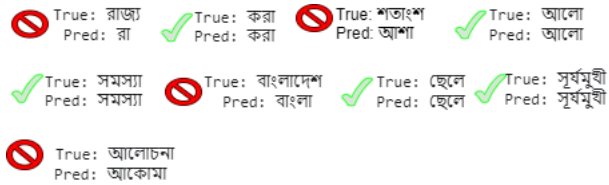

Fig. 9: Prediction on test dataset from the model trained with overlapped dataset

## V. QUALITATIVE STUDY AND RESULT

After generating the synthetic data, we verified it with a group of people. We picked a small sample of generated word images from both the overlapped and non-overlapped datasets and separated them into another folder for verification. A total of 30 participants was arranged for the qualitative study. The participants we chose have good knowledge of recognizing Bangla words. All of them were undergraduate students with an engineering background in a private university in Bangladesh. The study was conducted in two phases–1. they were assigned to give their opinion on the non-overlapped dataset, and 2. they were assigned to give their opinion on the overlapped dataset.

The participants were asked to label the images and vote on three criteria: how realistic the images are, how difficult they are to recognize, and how easy they are to recognize.

The result on non-overlapped dataset is shown in Figure 10.

Fig. 10: Study on non-overlapped images

The figure shows all of the images are predicted accurately. In terms of voting criteria, all of the votes given as the images are realistic. Furthermore, we measured the difficulty level for recognizing the words since words with more than three characters become difficult to understand. However, we received mixed votes for the criteria of understanding level. Both the easy-to-understand and difficult-to-understand criteria receive almost equal votes. Images with more than two characters are challenging for a few people to understand. On the contrary, images with two characters did not receive any vote from difficult-to-understand criteria.

the result on the overlapped dataset is shown in Figure 11. The figure shows that most of the images are predicted accurately except the image-id 2. Most votes are given as " the images are realistic". In terms of the difficulty level and ease level, the votes received equally in both of the criteria. Some of the words become difficult to understand to recognize due to the overlapping operation.

## VI. DISCUSSIONS

We trained our BiLSTM-CTC model with both overlapped and non-overlapped data. After training both types of data, we evaluated the performance of the model using WER, accuracy, precision, recall, and f1-score on an unseen dataset. From the analysis with both types of data, we achieved the highest accuracy on the non-overlapped dataset. However, the model showed a poor result using the overlapped dataset. The reason is that our test dataset does not contain overlapping instances. The presence of noise can be another reason for performing poorly. Since the non-overlapped dataset does not contain noises, it performed well on similar types of datasets.

We also verified a sample of our large dataset from human knowledge. All of the participants we hired are educated and can recognize the Bangla word properly since they are Bangladeshi. From this verification, we observed that the non-overlapped words are easy to recognize for most of the participants. The words are also realistic to their opinion. However, many of the participants face difficulty in recognizing the overlapped words. They came to the point that those seem realistic but challenging to understand. In general, many handwritten words are difficult to understand since the style varies from person to person. Therefore, the difficulty level of understanding the words has a positive side for recognizing noisy datasets. The purpose of creating the overlapped data is to construct a dataset that becomes a bit challenging to understand, and the qualitative study satisfied the purpose. Therefore, Those words can be helpful for different purposes with the presence of noises.

| imag_id | image | Participants prediction | no of participants voted as natural handwritten word | no of participants voted as difficult to recognize the word | no of participants voted as easy to recognize the word |
|---|---|---|---|---|---|
| 1 | 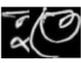 | Actual: হতে<br>pred: হতে | 29 | 7 | 23 |
| 2 | 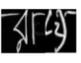 | Actual: রাজ্য<br>pred: রাদ্য | 27 | 30 | 0 |
| 3 | 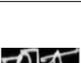 | Actual: কাক<br>pred: কাক | 30 | 0 | 30 |
| 4 | 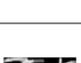 | Actual: দেশ<br>pred: দেশ | 27 | 26 | 4 |
| 5 | 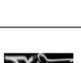 | Actual: নাম<br>pred: নাম | 30 | 3 | 27 |

Fig. 11: Study on overlapped images

## VII. CONCLUSION

We have proposed a novel way of generating handwritten words. We tried to solve issues like the lack of a large dataset. However, diverse and meaningful data make sure to recognize real-world data efficiently. Since the words were generated following the dictionary, it does not contain any noise that may cause the wrong learning process in the model. To the best of our knowledge, this work tried to bring an efficient and effective way which has not been done yet. However, a lot of work has been done on character-level text recognition, ensuring that most of the language has a sufficient handwritten character dataset but not a word-level handwritten dataset. Using our approach, generating a diverse and large dataset is simple and efficient. This paper is beneficial for recognizing handwritten text from the document for language that does not have a word-level handwritten dataset.